# GENETIC-ALGORITHM SEEDING OF IDIOTYPIC NETWORKS FOR MOBILE-ROBOT NAVIGATION

Amanda M. Whitbrook, Uwe Aickelin, Jonathan M. Garibaldi
*ASAP Research Group, School of Computer Science, University of Nottingham, Jubilee Campus, Wollaton Road, Nottingham, NG8 1BB, UK*
*amw@cs.nott.ac.uk, uxa@cs.nott.ac.uk, jmg@cs.nott.ac.uk*

Abstract: Robot-control designers have begun to exploit the properties of the human immune system in order to produce dynamic systems that can adapt to complex, varying, real-world tasks. Jerne's idiotypic-network theory has proved the most popular artificial-immune-system (AIS) method for incorporation into behaviour-based robotics, since idiotypic selection produces highly adaptive responses. However, previous efforts have mostly focused on evolving the network connections and have often worked with a single, pre-engineered set of behaviours, limiting variability. This paper describes a method for encoding behaviours as a variable set of attributes, and shows that when the encoding is used with a genetic algorithm (GA), multiple sets of diverse behaviours can develop naturally and rapidly, providing much greater scope for flexible behaviour-selection. The algorithm is tested extensively with a simulated e-puck robot that navigates around a maze by tracking colour. Results show that highly successful behaviour sets can be generated within about 25 minutes, and that much greater diversity can be obtained when multiple autonomous populations are used, rather than a single one.

## 1 INTRODUCTION

Short-term learning can be defined as the training that takes place over the lifetime of an individual, and long-term learning as that which evolves and develops as a species interacts with its environment and reproduces itself. The vertebrate immune system draws on both types since, at birth, an individual possesses a pool of antibodies that has evolved over the lifetime of the species; the repertoire also adapts and changes over the lifetime of the individual as the living body responds to invading antigens. Recently, researchers have been inspired by the learning and adaptive properties of the immune system when attempting to design effective robot-navigation systems. Many artificial-immune-system (AIS) methodologies adopt the analogy of antibodies as robot behaviours and antigens as environmental stimuli. Farmer's computational model (Farmer *et al.*, 1986) of Jerne's idiotypic-network theory (Jerne, 1974), which assumes this relation, has proved an extremely popular choice, since the antibody (behaviour) that best matches the invading antigen (current environment) is not necessarily selected for execution, producing a flexible and dynamic system.

The idiotypic architecture has produced some encouraging results, but has generally suffered from the same problems as previous approaches, as most designs have used small numbers of pre-engineered behaviours, limiting the self-discovery and learning properties of the schemes. This research aims to solve the problem by encoding behaviours as a set of variable attributes and using a genetic-algorithm (GA) to obtain diverse sets of antibodies for seeding the AIS. Here, the first phase of the design is described, i.e. the long-term phase that seeks to produce the initial pool of antibodies.

The long-term phase is carried out entirely in simulation so that it can execute as rapidly as possible by accelerating the simulations to maximum capacity. The population size is varied and, in addition, two different population models are considered, since it is imperative that an idiotypic system is able to select from a number of very diverse behaviours. In the first scheme there is only one population, but in the second, separate populations evolve in series but never interbreed. In each case the derived antibody-sets are scored in terms of diversity, solution quality, and how quickly they evolve.

The paper is arranged as follows. Section 2 shows how the vertebrate immune system depends on both short-term and long-term learning, and discusses how AIS has been used as a model for robotic controllers. It also highlights some of the problems with previous approaches to AIS robot-control and with evolutionary robotics in general. Section 3 describes the test environments and the problem used, and section 4 focuses on the architecture of the long-term phase including details of the GA. The experimental procedures are outlined in Section 5 and the results are presented and discussed in Section 6. Section 7 concludes the paper.

## 2 BACKGROUND AND MOTIVATION

Throughout the lifetime of an individual, the adaptive immune system learns to recognise antigens by building up high concentrations of antibodies that have proved useful in the past, and by eliminating those deemed redundant. This is a form of short-term learning. However, the antibody repertoire is not random at birth and the mechanism by which antibodies are replaced is not a random process. Antibodies are built from gene libraries that have evolved over the lifetime of the species. This demonstrates that the immune system depends on both short-term and long-term learning in order to achieve its goals.

When using the immune system as inspiration for robot controllers, many researchers opt to implement an idiotypic network based on Farmer's model of continuous antibody-concentration change. In this model the concentrations are not only dependent on the antigens, but also on the other antibodies present in the system, i.e. antibodies are suppressed and stimulated by each other as well as being stimulated by antigens. In theory this design permits great variability of robot behaviour since the antibodies model the different behaviours, and the complex dynamics of stimulation and suppression ensure that alternative antibodies are tried when the need arises (Whitbrook *et al.*, 2007). However, past work in this area has mostly focused on how the antibodies in the network should be connected and, for simplicity, has used a single set of pre-engineered behaviours for the antibodies, which limits the potential of the method. For example, Watanabe *et al.* (1998a, 1998b) use an idiotypic network to control a garbage-collecting robot, utilizing GAs to evolve their initial set of antibodies. The antibodies are composed of a precondition, a behaviour, and an idiotope part that defines antibody connection. However, the sets of possible behaviours and preconditions are fixed; the GA works simply by mixing and evolving different combinations with various parameters for the idiotope. Michelan and Von Zuben (2002) and Vargas *et al.* (2003) also use GAs to evolve the antibodies, but again only the idiotypic-network connections are derived. Krautmacher and Dilger (2004) apply the idiotypic method to robot navigation, but their emphasis is on the use of a variable set of antigens; they do not change or develop the initial set of handcrafted antibodies, as only the network links are evolved. Luh and Liu (2004) address target-finding using an idiotypic system, modelling their antibodies as steering directions. However, although many behaviours are technically possible since any angle can be selected, the method is limited because a behaviour is defined only as a steering angle and there is no scope for the development of more complex functions. Hart *et al.* (2003) update their network links dynamically using reinforcement learning, but use a skill hierarchy so that more complex tasks are achieved by building on basic ones, which are hand-designed at the start.

It is clear that the idiotypic AIS methodology holds great promise for providing a system that can adapt to change, but its potential has never been fully explored because of the limits imposed on the fundamental behaviour-set. This research aims to widen the scope of the idiotypic network by providing a technique that rapidly evolves simple, distinct behaviours in simulation. The behaviours can then be passed to a real robot as a form of intelligent initialization, i.e. a starting set of behaviours would be available for each known antigen, from which the idiotypic selection-mechanism could pick.

In addition, long-term learning in simulation coupled with an idiotypic AIS in the real world represents a novel combination for robot-control systems, and provides distinct advantages, not only for AIS initialization, but also for evolutionary robotics. In the past, much evolutionary work has been carried out serially on physical robots, which requires a long time for convergence and puts the robot and its environment at risk of damage. For example, Floreano and Mondada (1996) adopt this approach and report a convergence time of ten days. More recent evolutionary experiments with physical robots, for example Marocca and Floreano (2002), Hornby *et a*l. (2000), and Zykov *at al*. (2004) have produced reliable and robust systems, but have not overcome the problems of potential damage and slow, impractical convergence times. Evolving in parallel with a number of robots, (for example Watson *et al*. 1999) reduces the time required, but can still be extremely prohibitive in terms of time and logistics. Simulated robots provide a definite advantage in terms of speed of convergence, but the trade-off is the huge difference between the simulated and real domains (Brooks, 1992).

Systems that employ an evolutionary training period (long-term leaning phase) and some form of lifelong adaptation (short-term learning phase) have been used to try to address the problem of domain differences, for example by Nehmzow (2002). However, the long-term learning phase in Nehmzow's work uses physical robots evolved in parallel, which means that the method is slow and restricted to multi-agent tasks. Floreano and Urzelai (2000) evolve an adaptable neural controller that transfers to different environments and platforms, but use a single physical robot for the long-term phase. Keymeulen *et al.* (1998) run their long-term and short-term learning phases simultaneously, as the physical robot maps its environment at the same time as carrying out its goal-seeking task, thus creating the simulated world. They report the rapid evolution of adaptable and fit controllers, but these results apply only to simple, structured environments where the robot can always detect the coloured target, and the obstacles are few. For example, they observe the development of obstacle avoidance in five minutes, but this applies to an environment with only one obstacle, and the results imply that the real robot was unable to avoid the obstacle prior to this. Furthermore, only eight different types of motion are possible in their system. Walker *et al*. (2006) use a GA in the simulated long-term phase and an evolutionary strategy (ES) on the physical robot. They note improved performance when the long-term phase is implemented, and remark that the ES provides continued adaptation to the environment, but they deal with only five or 21 behaviour parameters in the GA, and do not state the duration of the long-term phase.

The method described here aims to capitalize on the fast convergence speeds that a simulator can achieve, but will also address the domain compatibility issues by validating and, if necessary, modifying all simulation-derived behaviours in the real world. This will be achieved by transferring the behaviours to an adaptive AIS that runs on a real robot. The method is hence entirely practical for real world situations, in terms of delivering a short training-period, safe starting-behaviours, and a fully-dynamic and adaptable system.

## 3 TEST ENVIRONMENT AND PROBLEM

The long-term phase requires accelerated simulations in order to produce the initial sets of antibodies as rapidly as possible. For this reason the Webots simulator (Michel, 2004) is selected as it is able to run simulations up to 600 times faster than real time, depending on computer power, graphics card, world design and the number and complexity of the robots used. The chosen robot is the e-puck (see Figure 1), since the Webots c++ environment natively supports it. It is a miniature mobile-robot equipped with a ring of eight noisy, nonlinear, infra-red (IR) sensors that can detect the presence of objects up to a distance of about 0.1 m. It also has a small frontal camera that receives the raw RGB values of the images in its field-of-view. Blob-finding software is created to translate this data into groups of like-coloured pixels (blobs).

The test problem used here consists of a virtual e-puck that must navigate around a building with three rooms (see Figures 2 and 3) by tracking blue markers painted on the walls. These markers are intended to guide the robot through the doors, which close automatically once the robot has passed through. The course is completed once the robot has crossed the finish-line in the third room. Its performance is measured according to how quickly it reaches the finish-line and how many times it collides with the walls or obstacles placed in the rooms. Two different test environments are used; World 1 (see Figure 2) has fewer obstacles and no other robots. World 2 (see Figure 3) contains more obstacles, and there is also a dummy wandering-robot in each room.

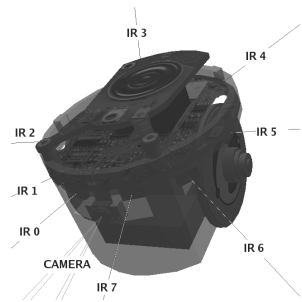

Figure 1: A simulated e-puck robot.

The simulations are run in fast mode (no graphics) with Webots version 5.1.10 using GNU/Linux 2.6.9 (CentOS distribution) with a Pentium 4 processor (clock speed 3.6 GHz). The graphics card used is an NVIDIA GeForce 7600GS, which affords average simulation speeds of approximately 200-times real-time for World 1 and 100-times real-time for World 2. The camera field-of-view is set at 0.3 radians, the pixel width and height at 15 and 3 pixels respectively and the speed unit for the wheels is set to 0.00683 radians/s.

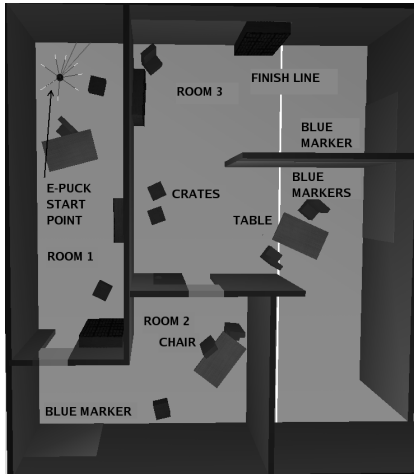

Figure 2: World 1 showing e-puck start point.

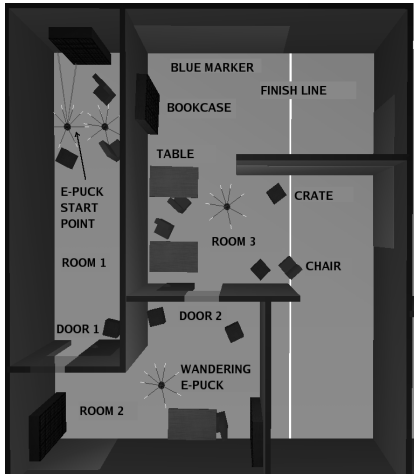

Figure 3: World 2 showing e-puck start-point, dummy-robot start-point and dummy-robot repositioning points.

## 4 SYSTEM ARCHITECTURE

### 4.1 Antigens and Antibodies

The antigens model the environmental information as perceived by the sensors. In this problem there are only two basic types of antigen, whether a door-marker is visible (a "marker" type antigen) and whether an obstacle is near (an "obstacle" type antigen), the latter taking priority over the former. An obstacle is detected if the IR sensor with the maximum reading $I_{max}$ has value $V_{max}$ equal to 250 or more. (N. B. The IR sensors are used in active mode where the readings correspond to the quantity of reflected light.) If no obstacles are detected then the perceived antigen is of type "marker" and there are two varieties, "marker seen" and "marker unseen", depending on whether appropriate-coloured pixel-clusters have been recognized by the blob-finding software. If an obstacle is detected then the antigen is of type "obstacle", i.e. the robot is no longer concerned with the status of the door-marker. The obstacle is classified in terms of both its distance from and its orientation toward the robot. The distance is "near" if $V_{max}$ is between 250 and 2400 and "collision" if $V_{max}$ is 2400 or more. The orientation is "right" if $I_{max}$ is sensor 0, 1 or 2, "rear" if it is 3 or 4 and "left" if it is 5, 6 or 7 (see Figure 1). There are thus eight possible antigens, which are assigned a code value 0–7, see Table 1.

Table 1: System antigens.

| Antigen Code | Antigen Type | Name |
|---|---|---|
| 0 | Marker | Marker unseen |
| 1 | Marker | Marker seen |
| 2 | Obstacle | Obstacle near right |
| 3 | Obstacle | Obstacle near rear |
| 4 | Obstacle | Obstacle near left |
| 5 | Obstacle | Collision right |
| 6 | Obstacle | Collision rear |
| 7 | Obstacle | Collision left |

The behaviours that form the core of the antibodies are encoded using a structure that has the attributes, type $T$, speed $S$, frequency of turn $F$, angle of turn $A$, direction of turn $D$, frequency of right turn $R_f$, angle of right turn $R_a$, and cumulative reinforcement-learning score $L$. There are six types of behaviour; wandering using either a left or right turn, wandering using both left and right turns, turning forwards, turning on the spot, turning backwards, and tracking the door-markers. The fusion of these basic behaviour-types with a number of different attributes that can take many values means that millions of different behaviours are possible. However, some behaviour types do not use a particular attribute and there are limits to the values that the attributes can take. These limits are carefully selected in order to strike a balance between reducing the size of the search space, which increases speed of convergence, and maintaining diversity, see Table 2.

## 4.2 System Structure

The control program uses the two-dimensional array of behaviours $B_{ij}$, $i = 0, \ldots, x-1$, $j = 0, \ldots, y-1$, where $x$ is the number of robots in the population ($x \geq 5$) and $y$ is the number of antigens, i.e. eight. When the program begins $i$ is equal to zero, and the array is initialized to null. The infra-red sensors are read every 192 milliseconds and the camera is read every 384 milliseconds, but only if no obstacles are found, as this increases computational efficiency.

Once an antigen code is determined, a behaviour or antibody is created to combat it by randomly choosing a behaviour type and its attribute values. For example, the behaviour WANDER_SINGLE (605, 50, 90, LEFT, NULL, NULL) may be created. This behaviour consists of travelling forwards with a speed of 605 Speed Units/s, but turning left 50% of the time by reducing the speed of the left wheel by 90%. If the antigen code is 7 then the $S$, $F$, $A$, $D$, $R_f$ and $R_a$ attributes of $B_{07}$ take the values 605, 50, 90, LEFT, NULL, NULL respectively. The action is executed and the sensor values are read again to determine the next antigen code. If the antigen has been encountered before, then the behaviour assigned previously is used, otherwise a new behaviour is created. The algorithm proceeds in this manner, creating new behaviours for antigens that have not been seen before and reusing the behaviours allotted to those that have.

However, the performance of the behaviours in dealing with the antigen they have been allocated to is constantly assessed using reinforcement learning (see section 4.4), so that poorly-matched behaviours can be replaced with newly-created ones when the need arises. Behaviours are also replaced if the antigen has not changed in any 60-second period, as this most likely means that the robot has not undergone any translational movement. The cumulative reinforcement-score of the previously used behaviour $L$ is adjusted after every sensor reading, and if it falls below the threshold value of -14 then replacement of the behaviour occurs. The control code also records the number of collisions $c_i$ for each robot in the population.

A separate supervisor-program is responsible for returning the virtual robot back to its start-point once it has passed the finish-line, for opening and closing the doors as necessary, and for repositioning the wandering dummy-robot, so that it is always in the same room as the mission robot. Another of the supervisor's functions is to assess the time taken $t_i$ to complete the task. Each robot is given 1250 seconds to reach the end-point; those that fail receive a 1000-second penalty if they did not pass through any doors. Reduced penalties of 750 and 500 seconds are awarded to failing robots that pass through one door and two doors respectively. When the whole population has completed the course, the relative-fitness $\mu_i$ of each individual is calculated. Since high values in terms of both $t_i$ and $c_i$ should yield a low relative-fitness, the following formula is used:

$$\mu_i = \frac{1}{(t_i + c_i)\sum_{k=0}^{x-1}(t_k + c_k)^{-1}}. \quad (1)$$

Table 2: System antibody types.

| No. | Description | $S$ Speed Units / s | | $F$ % of time | | $A$ % reduction in speed of one wheel | | $D$ Either left or right | | $R_f$ % of time | | $R_a$ % reduction in right wheel-speed | |
|---|---|---|---|---|---|---|---|---|---|---|---|---|---|
| | | MIN | MAX | MIN | MAX | MIN | MAX | 1 | 2 | MIN | MAX | MIN | MAX |
| 0 | Wander single | 50 | 800 | 10 | 90 | 10 | 110 | L | R | - | - | - | - |
| 1 | Wander both | 50 | 800 | 10 | 90 | 10 | 110 | - | - | 10 | 90 | 10 | 110 |
| 2 | Forward turn | 50 | 800 | - | - | 20 | 200 | L | R | - | - | - | - |
| 3 | Static turn | 50 | 800 | - | - | 100 | 100 | L | R | - | - | - | - |
| 4 | Reverse turn | 500 | 800 | - | - | 20 | 200 | L | R | - | - | - | - |
| 5 | Track markers | 50 | 800 | - | - | 0 | 30 | - | - | - | - | - | - |

The five fittest robots in the population are selected, and their mean $t_n$, $c_n$ and absolute-fitness $f_n$ are calculated, where $n$ represents the generation number, and $f_n = t_n + c_n$. In addition, the value of $f_n$ is compared with that of the previous generation $f_{n-1}$ to assess rate-of-convergence. The genetic algorithm is complete when any of the four conditions shown in Table 3 are reached. These are selected in order to achieve fast convergence, but also to maintain a high solution quality. Once convergence is achieved the attribute values representing the behaviours of the five fittest robots are saved for seeding the AIS system. If there is no convergence then the GA proceeds as described in section 4.3.

Table 3: Stopping criteria.

|   | Criteria - World 1 | Criteria - World 2 |
|---|---|---|
| 1 | $n > 0$ AND $t_n < 400$ AND $c_n < 60$ AND $\|f_n - f_{n-1}\| < 0.1$ | $n > 0$ AND $t_n < 600$ AND $c_n < 90$ AND $\|f_n - f_{n-1}\| < 0.2$ |
| 2 | $n > 30$ | $n > 30$ |
| 3 | $t_n < 225$ AND $c_n < 35$ | $t_n < 400$ AND $c_n < 45$ |
| 4 | $n > 15$ AND $\|f_n - f_{n-1}\| < 0.1$ | $n > 15$ AND $\|f_n - f_{n-1}\| < 0.2$ |

Note that when adopting the scenario of five separate populations that never interbreed, the five robots that are assessed for convergence are the single fittest from each of the autonomous populations. In this case, convergence is dependent upon the single best $t_n$, $c_n$ and $f_n$ values. The final five robots that pass their behaviours to the AIS system are the single fittest from each population after convergence.

### 4.3 The Genetic Algorithm

Two different parent robots are selected through the roulette-wheel method and each of the $x$ pairs interbreeds to create $x$ child robots. This process is concerned with assigning behaviour attribute-values to each of the $x$ new robots for each of the $y$ antigens in the system. It can take the form of complete antibody replacement, attribute-value mutation, adoption of the attribute values of only one parent or crossover from both parents.

Complete antibody replacement occurs according to the prescribed mutation rate ε. Here, a completely new random behaviour is assigned to the child robot for the particular antigen, i.e. both the parent behaviours are ignored.

Crossover is used when there has been no complete replacement, and the method used depends on whether the parent behaviours are of the same type. If the types are different then the child adopts the complete set of attribute values of one parent only, which is selected at random. If the types are the same, then crossover can occur by taking the averages of the two parent values, by randomly selecting a parent value, or by taking an equal number from each parent according to a number of set patterns. In these cases, the type of crossover is determined randomly with equal probability. The purpose behind this approach is to attempt to replicate nature, where the offspring of the same two parents may differ considerably each time they reproduce.

Mutation of an attribute value may also take place according to the mutation rate ε, provided that complete replacement has not already occurred. Here, the individual attribute-values (all except $D$) of a child robot may be increased or decreased by between 20% and 50%, but must remain within the prescribed limits shown in Table 2.

### 4.4 Reinforcement Learning

Reinforcement learning is used in order to accelerate the speed of the GA's convergence. It can be thought of as microcosmic short-term learning within the long-term learning cycle. The reinforcement works by comparing the current and previous antibody codes, see Table 4. Ten points are awarded for every positive change in the environment, and ten are deducted for each negative change. For example, 20 points are awarded if the antigen code changes from an "obstacle" type to "marker seen", because the robot has moved away from an obstacle as well as gaining or keeping sight of a door-marker.

Table 4: Reinforcement scores.

| Antigen code | | Reinforcement status (score) |
|---|---|---|
| Old | New | |
| 0 | 0 | Neutral (0) |
| 1 | 0 | Penalize - Lost sight of marker (-10) |
| 2-7 | 0 | Reward - Avoided obstacle (10) |
| 0 | 1 | Reward - Found marker (10) |
| 1 | 1 | Reward – Kept sight of marker (Score depends on orientation of marker with respect to robot) |
| 2-7 | 1 | Reward - Avoided obstacle and gained or kept sight of marker (20) |
| 0 | 2-7 | Neutral (0) |
| 1 | 2-7 | Neutral (0) |
| 2-7 | 2-7 | Reward or Penalize (Score depends on several factors) |

In the case where the antigen code remains at 1 (a door-marker is kept in sight), the score awarded depends upon how the orientation of the marker has moved with respect to the robot. In addition, when an obstacle is detected both in the current and previous iteration, then the score awarded depends upon several factors, including changes in the position of $I_{max}$ and in the reading $V_{max}$, the current and previous distance-type ("collision" or "near") and the tallies of consecutive "nears" and "collisions".

## 5 EXPERIMENTAL PROCEDURES

### 5.1 General Procedures

The GA is run in Worlds 1 and 2 using single populations of 25, 40, and 50 robots, and using five autonomous populations of five, eight, and ten. A mutation rate $\varepsilon$ of 5% is used throughout, as previous trials have shown that this provides a good compromise between fast convergence, high diversity and good solution-quality. Solution quality is measured as $q = (t + 8c)/2$, as this allows equal weighting for the number of collisions. For each scenario, ten repeats are performed and the means of the program execution time $\tau$, solution quality $q$, and diversity in type $Z_t$ and speed $Z_s$ are recorded. The mean solution-quality is also noted when 240 repeats are performed in each world using a hand-designed controller. This shows how well the GA-derived solutions compare with an engineered system and provides an indication of problem difficulty. Two–tailed standard t-tests are conducted on the result sets, and differences are accepted as significant at the 99% level only.

In World 2 the stopping criteria is relaxed in order to improve convergence speed, see Table 3. This is necessary since there are more obstacles and moving robots to navigate around, which means that completion time is affected.

### 5.2 Measuring Diversity

Diversity is measured using the type $T$ and the speed $S$ attributes of each of the final antibodies passed to the AIS system, since these are the only action-controlling attributes that are common to all antibodies. The antibodies are arranged into $y$ groups of five ($y$ is the number of antigens) and each group is assessed by comparison of each member with the others, i.e. ten pair-wise comparisons are made in each group. A point is awarded for each comparison if the attribute values are different; if they are the same no points are awarded. For example, the set of behaviour types [1 3 4 4 1] has two pair-wise comparisons with the same value, so eight points are given. Table 5 summarizes possible attribute-value combinations and the result of conducting the pair-wise comparisons on them.

Table 5: Diversity scores.

| Attribute-value status | Points | Expected: | |
|---|---|---|---|
| | | Frequency for $T$ | Score for $T$ |
| All five different | 10 | 9.26 | 0.926 |
| One repeat of two | 9 | 46.30 | 4.167 |
| Two repeats of two | 8 | 23.15 | 1.852 |
| One repeat of three | 7 | 15.43 | 1.080 |
| Two repeats, one of two, one of three | 6 | 3.86 | 0.231 |
| One repeat of four | 4 | 1.93 | 0.077 |
| All five the same | 0 | 0.08 | 0.000 |
| **Total** | | **100.00** | **8.333** |

The $y$ individual diversity-scores for each of $T$ and $S$ are summed and divided by $\sigma y$ to yield a diversity score for each attribute. Here $\sigma$ is the expected diversity-score for a large number of randomly-selected sets of five antibodies. This is approximately 8.333 for $T$ (see Table 5) and 10.000 for $S$. It is lower for $T$ since there are only six behaviours to select from, whereas the speed is selected from 751 possible values, so one would expect a random selection of five to yield a different value each time. The adjustment effectively means that a random selection yields a diversity of 1 for both $S$ and $T$. The diversity calculation is given by:

$$Z = \frac{\sum_{i=1}^{y} z_i}{\sigma y}, \qquad (2)$$

where $Z$ represents the overall diversity-score and $z$ represents the individual score awarded to each antigen.

## 6 RESULTS AND DISCUSSION

Table 6 presents mean $\tau$, $q$, $Z_t$, and $Z_s$ values in World 1, and Table 7 summarises the significant difference levels when comparing single and multiple populations. The schemes that are compared use the same number of robots, for example a single population of 25 is compared with five populations of five. In addition, the smallest and largest population sizes are compared for both single and multiple populations.

Table 6: World 1 means.

| Pop. size | $\tau$ (s) | $q$ | $Z_t$ (%) | $Z_s$ (%) |
|---|---|---|---|---|
| 25 | 417 | 220 | 40 | 86 |
| 40 | 530 | 216 | 53 | 95 |
| 50 | 811 | 191 | 49 | 90 |
| 5 x 5 | 508 | 155 | 55 | 100 |
| 5 x 8 | 590 | 146 | 54 | 100 |
| 5 x 10 | 628 | 144 | 58 | 100 |

Table 7: World 1 significant differences.

| Comparison | | $\tau$ (s) | $q$ | $Z_t$ (%) | $Z_s$ (%) |
|---|---|---|---|---|---|
| 25 | 5 x 5 | 77.40 | **99.94** | **99.90** | **99.99** |
| 40 | 5 x 8 | 72.58 | **99.97** | 43.07 | **99.97** |
| 50 | 5 x 10 | 97.13 | **99.80** | 98.36 | **99.96** |
| 25 | 50 | **99.99** | 91.76 | 96.10 | 75.81 |
| 5 x 5 | 5 x 10 | 88.41 | 58.13 | 60.40 | 00.00 |

The tables show that there are no significant differences between controller run-times when comparing the single and multiple populations. Type diversity is consistently higher for the multiple populations, but only significantly higher when comparing a single population of 25 with five populations of five. However, solution quality and speed diversity are significantly better for the multiple populations in all three cases. Multiple populations always demonstrate a speed diversity of 100%, indicating that the final-selected genes are completely unrelated to each other, as expected. In contrast, single-population speed-diversity never reaches 100% as there are always repeated genes in the final-selected robots. Evidence from previous experiments with single populations of five, ten and 20 suggests that the level of gene duplication decreases as the single population size increases. This explains the lower $Z_t$ and $Z_s$ values for a population of 25 robots. However, when comparing the results from single populations of 25 with 50, the only significant difference is in the run-time, with 25-robot populations running much faster. This is intuitive, since fewer robots must complete the course for every generation. There are no significant differences when comparing five-robot and ten-robot multiple populations. Run-times may be comparable here because the course has to be completed fewer times for the smaller population, but it requires more generations for convergence since there seems to be a reduced probability of producing successful robots.

In all cases, mean type-diversity ratings never reach 100%, yet mean speed-diversity is always 100% in the multiple populations, which shows there are no repeated genes. The reduced type-diversity ratings must therefore occur because the types are not randomly selected but chosen in a more intelligent way. The relatively small number of types (six) means that intelligent selection reduces the type diversity, whereas speed diversity is unaffected because there are many potentially-good speeds to choose from and convergence is rapid. It is likely that both intelligent selection and repeated genes decrease the type-diversity scores for the single populations, but in the multiple populations, the phenomenon is caused by intelligent selection only.

The hand-designed controller demonstrates a mean solution-quality of 336. (The scores from the 49 robots that failed to complete the course are not counted.) This is significantly worse than all of the multiple populations, but not significantly different to the single populations, although single-population quality scores are considerably better. The multiple populations may have an advantage over the single populations in terms of solution quality because, for each population, they require a fast time and few collisions for only one member in order to meet the convergence criteria. The single-population case demands good mean-scores from five robots.

Table 8 presents the significant difference levels when comparing the results from World 1 with those from World 2. There is a significant difference in run-time in every case, which is not surprising because the GAs in World 2 take, on average, 2.25 times as long to converge. This is partly due to the World 2 simulations running only half as fast as those in World 1 (because there are two robots to control) and partly because the problem is harder to solve. There are also significant differences in solution quality for 50-robot single populations and all the multiple populations, with World 2 producing the lower-quality solutions. (The 25-robot and 30-robot populations are almost significant.) This difference is due to the less-stringent convergence criteria in World 2. There are no significant differences in type diversity or speed diversity between the two worlds.

Tables 9 and 10 summarise the same data as Tables 6 and 7, but for World 2. When comparing single with multiple populations, the results reveal a similar pattern to World 1 in terms of run-time, type-diversity and speed-diversity, i.e., there are no significant differences between run-times, although they are slightly higher for the multiple populations. Type diversity is consistently better for the multiple populations, but only significantly higher when comparing a single population of 25 with a five-robot multiple population. Speed diversity is consistently significantly higher for the multiple populations, with all multiple populations producing 100% diversity. However, unlike World 1, there are no significant differences in solution quality, although the figures for the multiple populations are better in each case.

Table 8: World 1 compared with World 2.

| Pop. size | τ (s) | q | $Z_t$ (%) | $Z_s$ (%) |
|---|---|---|---|---|
| 25 | **99.99** | 97.61 | 50.34 | 11.19 |
| 40 | **99.99** | 96.61 | 24.70 | 84.41 |
| 50 | **99.93** | 99.97 | 80.09 | 87.67 |
| 5 x 5 | **99.99** | **99.99** | 57.16 | 66.94 |
| 5 x 8 | **100.00** | **99.99** | 14.38 | 0.00 |
| 5 x 10 | **100.00** | 99.86 | 27.51 | 66.94 |

Table 9: World 2 means.

| Pop. size | τ (s) | q | $Z_t$ (%) | $Z_s$ (%) |
|---|---|---|---|---|
| 25 | 972 | 314 | 37 | 85 |
| 40 | 1292 | 266 | 51 | 89 |
| 50 | 1414 | 250 | 56 | 94 |
| 5 x 5 | 1211 | 258 | 58 | 100 |
| 5 x 8 | 1325 | 225 | 55 | 100 |
| 5 x 10 | 1498 | 208 | 57 | 100 |

Table 10: World 2 significant differences.

| Comparison | | τ (s) | q | $Z_t$ (%) | $Z_s$ (%) |
|---|---|---|---|---|---|
| 25 | 5 x 5 | 88.47 | 84.51 | **99.96** | **99.63** |
| 40 | 5 x 8 | 20.91 | 94.09 | 61.19 | **99.28** |
| 50 | 5 x 10 | 40.78 | 97.31 | 18.34 | **99.87** |
| 25 | 50 | 98.79 | 90.17 | **99.50** | 93.43 |
| 5 x 5 | 5 x 10 | 94.36 | 97.97 | 22.16 | 00.00 |

When comparing the results from single populations of 25 robots with 50 robots, the only significant difference is in type diversity, with 50-robot populations producing more diverse sets of behaviour type. Since type diversity is also significantly higher in the five-robot multiple populations, this suggests there may be a threshold single population size, below which single populations are significantly less diverse in behaviour-type than their multiple-population counterparts. There are no significant differences when comparing five-robot and ten-robot multiple populations, although the higher solution-quality for ten robots almost reaches significance.

In World 2 the hand-designed controller produces a mean solution-quality of 623 (not counting the results from the 109 robots that failed to complete the course). The performance is significantly worse than the GA-derived solutions from both the single and multiple populations in World 2.

# 7 CONCLUSIONS AND FUTURE WORK

This paper has described a GA method for intelligently seeding an idiotypic-AIS robot control-system, i.e. it has shown how to prepare an initial set of antibodies for each antigen in the environment. Experiments with static and dynamic worlds have produced solution-sets with significantly better mean solution-quality than a hand-designed controller, and the system has been able to deliver the starting antibodies within about ten minutes in the static world, and within about 25 minutes in the dynamic world. These are fast results compared with GAs that have used physical robots and reported convergence in terms of number of days rather than minutes. The method hence provides a practical training-period when considering real-world tasks.

The resulting antibody sets have also been tested for quality and diversity, and it has been shown that significantly higher antibody diversity can be obtained when a number of autonomous populations are used, rather than a single one. For sets of five populations, the mean diversity of antibody speed is 100%, and one can run the genetic algorithm without significantly increasing the convergence time or reducing solution quality. In fact, for simpler problems, multiple populations may help to improve solution quality. Results have also shown that the diversity ratings are not affected by the difficulty of the problem.

The potential of the method to create high behaviour-diversity augurs well for the next stage of the research, which is transference to a real robot running an AIS. This part of the work will investigate how the idiotypic-selection process should choose between the available solutions, and how antibodies should be replaced within the system when they have not proved useful. The work will also examine how closely the simulated-world needs to resemble the real-world in order that the initial solutions are of benefit.